\documentclass[conference]{IEEEtran}
\IEEEoverridecommandlockouts
\usepackage[compress]{cite}
\usepackage{amsmath,amssymb,amsfonts}
\usepackage{booktabs}
\usepackage{multirow}
\usepackage{algorithmic}
\usepackage{graphicx}
\usepackage{textcomp}
\usepackage[dvipsnames]{xcolor}
\usepackage{xspace}
\usepackage{mathtools}
\usepackage[caption=false]{subfig}
\usepackage[acronym]{glossaries}
\usepackage{balance}
\usepackage[normalem]{ulem}
\usepackage[colorlinks=true,allcolors=black]{hyperref}

\makeatletter
\DeclareRobustCommand\onedot{\futurelet\@let@token\@onedot}
\def\@onedot{\ifx\@let@token.\else.\null\fi\xspace}

\def\eg{e.g\onedot}

\def\etal{et.~al\onedot}
\makeatother

\newacronymstyle{long-short-br}
{%
  \GlsUseAcrEntryDispStyle{long-short}%
}%
{%
  \GlsUseAcrStyleDefs{long-short}%
}
\setacronymstyle{long-short-br}

\def\BibTeX{{\rm B\kern-.05em{\sc i\kern-.025em b}\kern-.08em
    T\kern-.1667em\lower.7ex\hbox{E}\kern-.125emX}}
\begin{document}

\title{Deep Learning for Image-based Automatic Dial Meter Reading: Dataset and Baselines}

\author{
\IEEEauthorblockN{Gabriel Salomon\IEEEauthorrefmark{1}, Rayson Laroca\IEEEauthorrefmark{1} and David Menotti\IEEEauthorrefmark{1}}
\IEEEauthorblockA{
\IEEEauthorrefmark{1}Department of Informatics, Federal University of Paran\'a (UFPR), Curitiba, PR, Brazil \\
\, Email: {\{gsaniceto, rblsantos, menotti\}}@inf.ufpr.br}
}

\maketitle

\newacronym{amr}{AMR}{Automatic Meter Reading}
\newacronym{copel}{Copel}{Energy Company of Paran\'a}
\newacronym{cnn}{CNN}{Convolutional Neural Network}
\newacronym{fcsrn}{FCSRN}{Fully Convolutional Sequence Recognition Network}
\newacronym{fps}{FPS}{frames per second}
\newacronym{ht}{HT}{Hough Transform}
\newacronym{hct}{HCT}{Hough Circle Transform}
\newacronym{hlt}{HLT}{Hough Line Transform}
\newacronym{ssd}{SSD}{Single Shot MultiBox Detector}
\newacronym{rnn}{RNN}{Recurrent Neural Networks}
\newacronym{map}{mAP}{mean Average Precision}
\newacronym{mlp}{MLP}{Multilayer Perceptron}
\newacronym{mser}{MSER}{Maximally Stable Extremal Regions}
\newacronym{hog}{HOG}{Histogram of Oriented Gradients}
\newacronym{roi}{ROI}{Region of Interest}
\newacronym{svm}{SVM}{Support Vector Machine}
\newacronym{ocr}{OCR}{Optical Character Recognition}
\newacronym{rpn}{RPN}{Region Proposal Network}
\newacronym{sift}{SIFT}{Scale-Invariant Feature Transform}
\newacronym{iou}{IoU}{Intersection over Union}

\newcommand{\faster}{Faster R-CNN\xspace}
\newcommand{\mask}{Mask R-CNN\xspace}
\newcommand{\dataset}{UFPR-ADMR\xspace}
\newcommand{\ufpramr}{UFPR-AMR\xspace}
\newcommand{\distance}{Levenshtein\xspace}
\begin{abstract}
Smart meters enable remote and automatic electricity, water and gas consumption reading and are being widely deployed in developed countries.
Nonetheless, there is still a huge number of non-smart meters in operation.
Image-based \gls*{amr} focuses on dealing with this type of meter readings.
We estimate that the \gls*{copel}, in Brazil, performs more than 850,000 readings of dial meters per month. 
Those meters are the focus of this work.
Our main contributions are: (i)~a public real-world dial meter dataset (shared upon request) called~\dataset; (ii)~a deep learning-based recognition baseline on the proposed dataset; and (iii)~a detailed error analysis of the main issues present in \gls*{amr} for dial meters. 
To the best of our knowledge, this is the first work to introduce deep learning approaches to multi-dial meter reading, and perform experiments on unconstrained images.
We achieved a 100.0\% F1-score on the dial detection stage with both \faster and YOLO, while the recognition rates reached 93.6\% for dials and 75.25\% for meters using \faster~(ResNext-101).
\end{abstract}


\section{Introduction}
\label{sec:introduction}
\glsresetall

Measuring residential energy consumption is known to be a laborious task~\cite{vanetti2013gas,gallo2015robust,li2019light}.
Although smart meters are gradually replacing old meters, there are still many old mechanical meters in operation around the world since their replacement is time-consuming and costly.
In many regions, such as remote areas and developing nations, manual on-site readings are still prevalent~\cite{laroca2019convolutional}.
Even in developed countries, replacements are still far for complete. 
For example, in the end of 2018, there were still more than $26$~million non-automatic meters in the United~States~\cite{b0}.

In the literature, \gls*{amr} is usually associated with digital and smart meters~\cite{kabalci2016survey}.
In this work, we use this designation exclusively for image-based automatic readings.
\gls*{amr} allows the employees of the service company (electricity/gas/water) or, preferably, the consumers themselves to capture meter images using a mobile device, which is cheaper and more feasible than manual on-site reading, and easier to deploy -- in the short/medium term -- than the replacement of old~meters.

\begin{figure}[!ht]
    \centering
    
    \vspace{-2mm}
    
    \subfloat[cyclometer display]{
        \includegraphics[height=3.5cm]{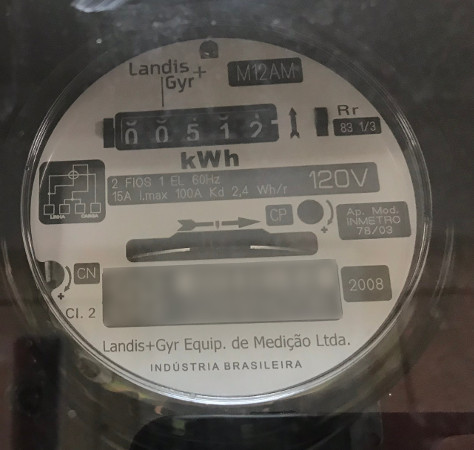}
    }
    \subfloat[dial display]{
        \includegraphics[height=3.5cm]{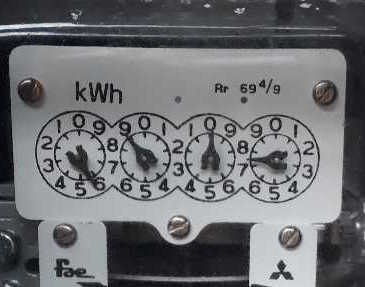}
     }
    
    \vspace{-1mm}
     
    \subfloat[electronic display]{
        \includegraphics[height=3.5cm]{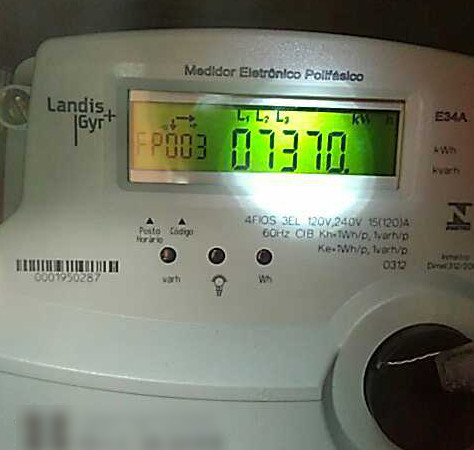}
     }
    \subfloat[smart meter]{
        \includegraphics[height=3.5cm]{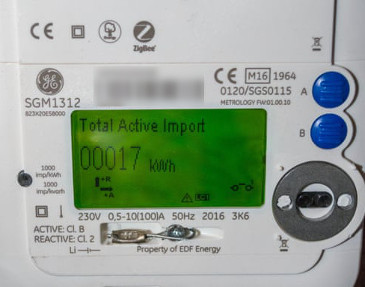}
     }
    \vspace{-0.5mm}
    \caption{The most common types of energy meters.}
    \label{fig:example}
\end{figure}

There are two main categories of residential energy meters~\cite{ausgrid, callmepower}: (i)~analog (with cyclometer and dial displays) and (ii)~digital (with electronic display and smart meters), as shown in Fig.~\ref{fig:example}.
This work focuses on dial meters since, although there are numerous dial meters in operation, there are still many open challenges in this context (as detailed~further).

The \gls*{copel}~\cite{copel} measures electricity consumption in more than $4$ million consuming units (i.e., meters) per month in the Brazilian state of~Paraná. 
From the images they provided us (see Section~\ref{sec:dataset}), we estimate that $21$\% of those devices are dial meters, resulting in more than $840{,}000$ dial meter readings carried out every~month.

Most of the dial meter recognition literature is focused on industrial applications, \eg, pressure meters~\cite{liu2019, zheng2017, huang2019}, voltmeter~\cite{jiale2011} and ammeter~\cite{fang2019}. 
As the meters are generally fixed and indoors, the image quality is strictly controlled.

Although in some cases the conditions are indeed realistic, they are not as unconstrained as in images obtained in outdoor environments, with challenging conditions,~\eg, severe lighting conditions (low light, glares, uneven illumination, reflections, etc.), dirt in the region of interest, and taken at a distance.
In addition, most approaches are based on handcrafted features~\cite{tang2015, zheng2017}, and were evaluated exclusively on private datasets~\cite{vega2013,tang2015,zheng2017,fang2019,liu2019}.
To the best of our knowledge, there are no public datasets containing dial meter images in the~literature.

Taking into account the above discussions, we introduce a real-world fully-labeled dataset (shared upon request) containing $2{,}000$ meter images, acquired in unconstrained scenarios by \gls*{copel} employees, with $9{,}097$ individual dials and a well-defined evaluation protocol to assist the development and assessment of new approaches for this~task\footnote{The \dataset dataset is \textbf{publicly available} (but upon request) to the research community
at \textit{\href{http://web.inf.ufpr.br/vri/databases/ufpr-admr/}{web.inf.ufpr.br/vri/databases/ufpr-admr/}}.}.
In addition, we conducted experiments using deep learning models in our dataset images to serve as baselines for future work, investigating problems related to dial meter reading and providing guidance for further research through a detailed quantitative and qualitative error~analysis.

The remainder of this work is organized as follows.
In Section~\ref{sec:related}, we discuss approaches designed for \gls*{amr} as well as deep learning techniques. 
The proposed dataset is described in Section~\ref{sec:dataset}. 
Section~\ref{sec:approaches} presents the evaluated deep learning-based approaches for automatic reading of dial meters, while the results (with a detailed error analysis) are reported in Section~\ref{sec:results}.
Lastly, in Section~\ref{sec:conclusions}, we state the conclusions.
\section{Related Work}
\label{sec:related}

There are many works in the literature that dealt with \gls*{amr}.
Most of them focus on the recognition of cyclometers and digital meters using \gls*{ocr} methods.
Recently, deep learning approaches have received great attention in this context~\cite{gomez2018cutting,huang2019,laroca2019convolutional,liu2019,tsai2019,yang2019}.
Dial meter recognition research, on the other hand, is more scarce.
Most methods focus on gauges for industrial application~\cite{jiale2011, zhang2016, zheng2017,fang2019, he2019, huang2019, liu2019}.
Although gauges may look similar to energy dial meters, they usually only contain a single dial and one type of dial template, and the image conditions tend to be much more controlled in terms of lighting, dirt, and image quality.
In this section, we describe some relevant works on \gls*{amr} as well as state-of-the-art deep learning approaches for object detection and recognition~\cite{he2016deep, simonyan2014very, ren2015faster, redmon2016yolo, redmon2018yolov3}.

\subsection{Digit-based Meter Reading}

Gallo~\etal~\cite{gallo2015robust} proposed a method that uses \gls*{mlp} to locate the \gls*{roi} of the meters (also denoted as counter region~\cite{vanetti2013gas, gallo2015robust, laroca2019convolutional}), \gls*{mser} to segment the digits, \gls*{hog} for feature extraction, and \gls*{svm} for digit~recognition. 

Nodari and Gallo~\cite{nodari2011multi} proposed a method named MultiNOD for gas cyclometers reading. It consists of a neural network tree, sharing and resizing features to perform counter detection and digit segmentation.
The digit recognition stage was handled using Tesseract.
This approach was later improved in~\cite{vanetti2013gas}, with the addition of a Fourier analysis applied to the segmented image, in order to avoid false positives.
Finally, \gls*{svm} was employed for digit classification.

Tsai~\etal~\cite{tsai2019} employed \gls*{ssd}~\cite{liu2016ssd}, a deep learning object detector, to locate the counter region in energy meters.
The authors reported an accuracy rate of $100$\% on their experiments, but did not address the recognition~stage.

Yang~\etal~\cite{yang2019} proposed a \gls*{fcsrn} for water meter analog digit reading, with a novel loss function entitled Augmented Loss~(AugLoss). AugLoss addresses the ``middle-state'' that can occur when the digit accumulator is changing from one display digit to the next one, usually outputting the old displayed digit.
Their approach outperformed \gls*{rnn} and attention-based models on the task of sequence~recognition, but the experiments were made in controlled images, with cropped and aligned meters. 

Gómez~\etal~\cite{gomez2018cutting} introduced a segmentation-free approach to perform meter reading. 
They trained a \gls*{cnn} to yield readings directly from the input images, without the need to detect the counter region. 
Although their approach has achieved promising results, the authors used a private dataset in the experiments, and only compared their method with traditional algorithms that rely on handcrafted~features, which are easily affected by noise and may not be robust to images acquired in adverse conditions~\cite{laroca2019convolutional}.

Laroca \etal~\cite{laroca2019convolutional} designed a two-stage approach for \gls*{amr}.
The Fast-YOLOv2 model~\cite{redmon2017yolo9000} was employed for counter detection and three \gls*{cnn}-based models were evaluated in the counter recognition stage.
The authors considerably improved their recognition results when balancing the training set in terms of digit classes through data augmentation techniques.

\subsection{Dial Meter Reading}

Tang~\etal~\cite{tang2015} proposed a complete framework for dial energy meter reading based on binarization, line intersection, and morphological operations.
Despite being an interesting approach, the dataset used in the experiments was not published, and the images were obtained in a controlled~environment.

In~\cite{vega2013}, the authors also employed handcrafted features for dial recognition.
In addition to binarization and line intersection, the counter region was detected using \gls*{sift} features.
Their method was evaluated on a private dataset containing only $141$ images taken in a controlled environment.

The following approaches dealt only with single-dial meters (commonly known as gauges) and not with energy meters.
Although the problems are similar, there is a fundamental difference: a small error in a multi-dial meter can result in a completely wrong measurement (especially if the error occurs in recognizing the most significant~dials).
Such a fact needs to be taken into account when evaluating recognition~methods.

Several approaches explored handcrafted features, such as \gls*{ht}, in order to locate the dials~\cite{jiale2011, zhang2016, zheng2017}.
The steps in such works are very similar: image binarization on the preprocessing stage, \gls*{hct} for dial location, and pointer angle detection using \gls*{hlt} or similar~methods.
These approaches generally work well in constrained environments, but may not be suitable for real-world outdoor scenarios with uneven lighting and the presence of~noise.

\mask was proposed for pointer recognition in~\cite{fang2019, he2019}.
Fang~\etal~\cite{fang2019} used it to find reference key points and the pointer in a gauge scale marks, while He~\etal~\cite{he2019} focused on segmenting the meter dial and pointer.
In both works, the angle between the pointer and the dial was explored to retrieve the reading.
The datasets used in the experiments were not provided in both works.

Region-based Fully Convolutional Networks (R-FCNs) were used for meter detection~\cite{huang2019}.
Although the authors used deep learning for detection, the meter reading was performed with handcrafted methods such as binarization, line detection, and skeleton extraction.
Liu~\etal~\cite{liu2019} evaluated Fast R-CNN, \faster, YOLO and \gls*{ssd} for meter detection and concluded that even though \faster outperforms the others, YOLO is the fastest. 
Nevertheless, the recognition was performed by a handcrafted method (i.e., \gls*{ht}) and the images used have not been made publicly~available.

\subsection{Deep Learning Methods}
ResNet~\cite{he2016deep} is one of the recent breakthroughs in deep networks.
The introduction of residual blocks enabled deeper network architectures while having fewer parameters than shallower networks, such as VGG19~\cite{simonyan2014very}.
ResNet also performs better and converges faster. The residual learning process introduces lower level features directly to higher abstraction layers, preserving information.
ResNet was later upgraded to ResNeXt~\cite{xie2017}. The main difference between them is the concept of ``cardinality''; instead of going deeper, ResNeXt uses a multi-branch architecture (cardinality refers to the number of branches used) to increase the transformations and achieve a higher representation power.
ResNet and ResNeXt can be employed for recognition (classification)~problems.

In order to detect the dials on each image, object detection deep networks will be explored.
\faster~\cite{ren2015faster} is a state-of-the-art approach that uses attention mechanisms and the sharing of convolutional features between the \gls*{rpn} and the detection network (originally VGG16) to enhance speed and accuracy.
First, the \gls*{rpn} generates region proposals that may contain known objects; then, the detection network evaluates the boundaries and classifies the~objects.

Redmon et al.~\cite{redmon2016yolo} proposed YOLO~(You Only Look Once), an object detector that focuses on an extreme speed/accuracy trade-off by dividing the input image into regions and predicting bounding boxes and probabilities for each region.
YOLOv2~\cite{redmon2017yolo9000}, an improved version of YOLO, adopts a series of concepts (e.g., anchor boxes, batch normalization, etc.) from existing works along with novel concepts to improve YOLO's accuracy while making it faster~\cite{liu2019deep}.
Similarly, Redmon and Farhadi~\cite{redmon2018yolov3} introduced YOLOv3 (the latest version of YOLO), which uses various tricks to improve training and increase performance, such as 
residual blocks, shortcut connections, and upsampling.
YOLO-based models have been successfully applied in several research~areas~\cite{laroca2019convolutional,laroca2019efficient,severo2018benchmark}.

\subsection{Datasets}

Most of the referred works do not provide a public dataset to enable a fair comparison of results.
There are a few publicly available meter datasets~\cite{yang2019,nodari2011multi, laroca2019convolutional}, however, none of them have images containing pointer-type meters, only digit-based ones.
As far as we know, there is no publicly available dataset containing images of dial~meters.
\section{The \dataset Dataset}
\label{sec:dataset}

We acquired the meter images from \gls*{copel}, a company of the Brazilian electricity sector that serves more than $4$~million consuming units per month~\cite{copel, laroca2019convolutional}.
The images of the meters were obtained at the consuming units by Copel employees using cell phone cameras (note that cell phones of many brands and models were~used).
All images had already been resized and compressed for storage, resulting in images of $640\times480$ or $480\times640$ pixels (depending on the orientation in which the image was taken).
To create the \dataset dataset, we selected $2{,}000$ images where it was possible for a human to recognize the correct reading of the meter, as the images were acquired in uncontrolled environments and it would not be possible to label the correct reading in many~cases.

\begin{figure}[!ht]
   \centering
    \includegraphics[width=\linewidth]{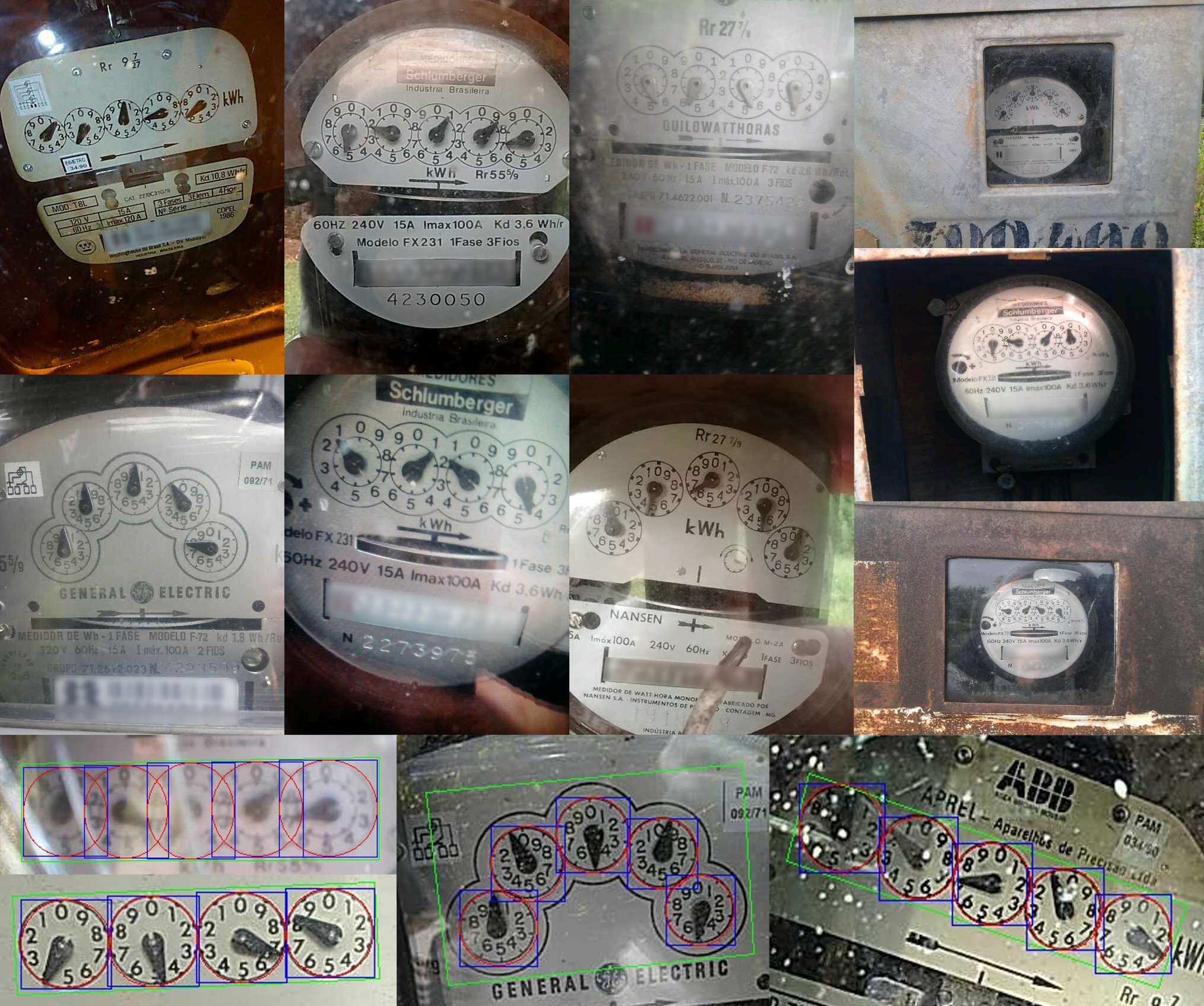}

    \caption{Examples of the images chosen for the dataset. In the bottom row, there are examples of the annotations provided for each image: in green the irregular surrounding quadrilateral, in blue the bounding boxes around the dials and in red the maximal ellipse contained in the bounding box. Note that the customer meter identification is blurred for keeping subject~privacy. 
    }
    \label{fig:examples-dataset}
\end{figure}

In each image, we manually labeled the position~($x$,~$y$) of each corner of an irregular quadrilateral that contains all the dials.
These corner annotations can be used to rectify the image patch containing the dials.
Fig.~\ref{fig:examples-dataset} shows some images selected for the dataset as well as illustrations of the~annotations.

All meters have $4$ or $5$ dials, being $903$ meters~($45$\%) with $4$ dials and $1{,}097$ meters~($55$\%) with $5$.
The values pointed on each dial have an almost uniform distribution of digits, having slightly more $0$s than other digits. 
Information about the dimensions of the meters and dials in the dataset are shown in Table~\ref{tab:sizedialsmeters}.
Note the great variability in the size of both meters and dials, for example, the smallest dial ($20\times29$ pixels) is almost $10$~times smaller than the largest one ($206\times201$~pixels).
 
 \begin{table}[!htb]   
    \caption{Statistics about the size of meters and individual dials.}
    \vspace{-1.5mm}
    \label{tab:sizedialsmeters}
    \resizebox{0.995\linewidth}{!}{%
    \, \begin{tabular}{cccccccc}
    \toprule
         & Min~(px) & Max~(px) & Mean~(px) & \multirow{2}{*}{Mean Area (px$^2$)}\\
     & W $\times$ H & W $\times$ H & W $\times$ H &  \\
    \midrule  
    Meters & $96$ $\times$ $37$  & $632$ $\times$ $336$ & $326$ $\times$ $121$ & $42{,}296$   \\
    Dials & $20$  $\times$ $29$  & $206$ $\times$ $201$  & $88$ $\times$ $86$ &  $8{,}328$ \\
    
    \bottomrule
    \end{tabular}
    }

\end{table}

\begin{figure}[!htb]
    \centering
    \hspace{-2.75mm}\includegraphics[width=0.99\linewidth]{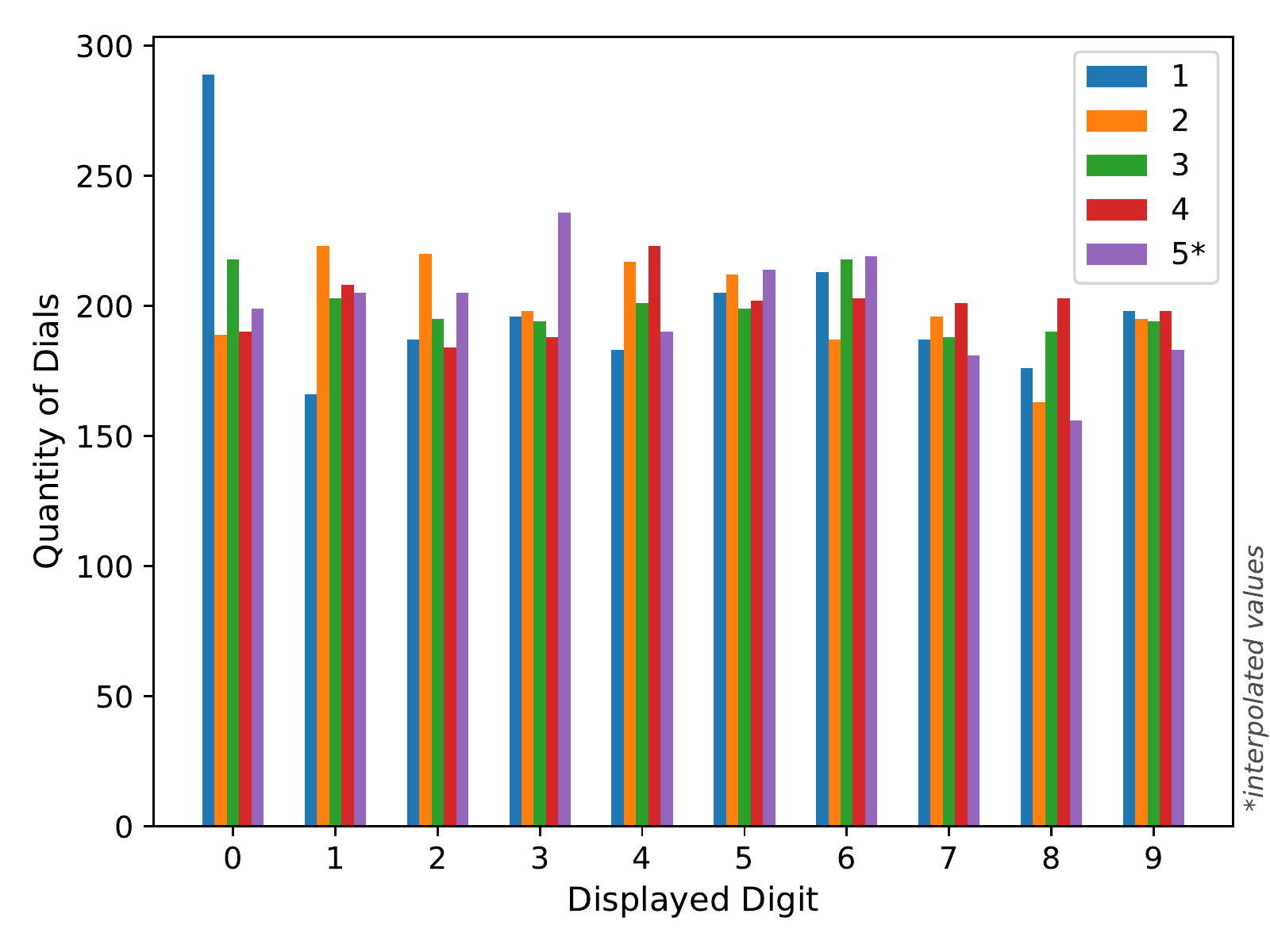}
    
    \vspace{-4.5mm}
    
    \caption{The distribution of digits according to the dial position on the meter. As $45$\% of the meters in the proposed dataset do not have a $5$th dial, the values of the $5$th dial quantities were interpolated proportionally for better~visualization.}
    \label{fig:graph_dist}
\end{figure}

Fig.~\ref{fig:graph_dist} illustrates the distribution of digits per dial.
The most prominent bar indicates that the most frequent digit in the first position is~$0$.
Nevertheless, it should be noted that the distribution is not as unbalanced as datasets with digit-based meter images, such as the \ufpramr dataset~\cite{laroca2019convolutional}, in which the number of $0$s in the first position is equal to the sum of $0$s in the other positions.
This is probably due to the fact that dial meters stopped being manufactured and deployed decades ago, which implies that each dial might have completed many cycles since the installation and may be indicating any~value.

Table~\ref{tab:freq} shows the frequency of digits in the \dataset dataset. 
Unlike datasets containing digit-based meters~\cite{yang2019, laroca2019convolutional}, which were manufactured/deployed more recently, the distribution of the digits is almost uniform across our~dataset.

 \begin{table}[!htb]   
    \centering
    \caption{Frequency distribution of digits in the \dataset dataset.}
    \label{tab:freq}
    
    \vspace{-1.5mm}
    \resizebox{.99\linewidth}{!}{
    \begin{tabular}{ccccccccccc}
    \toprule
    \multicolumn{10}{c}{Frequency / Digit Distribution}\\
    \midrule
    $0$ & $1$ & $2$ & $3$ & $4$ & $5$ & $6$ & $7$ & $8$ & $9$\\
    \midrule
     $996$ & $913$ & $899$ & $906$ & $929$ & $936$ & $942$ & $872$ & $818$ & $886$\\
    \bottomrule
    \end{tabular}
    }
\end{table}

\subsection{Challenges}

\begin{figure*}[!htb]
    \centering

    \subfloat[uneven lighting]{
        \includegraphics[width=.18\linewidth]{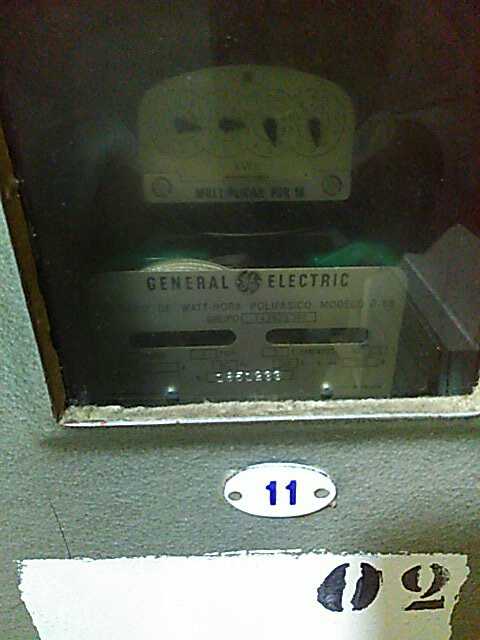}
        
    }
    \subfloat[blur]{
        \includegraphics[width=.18\linewidth]{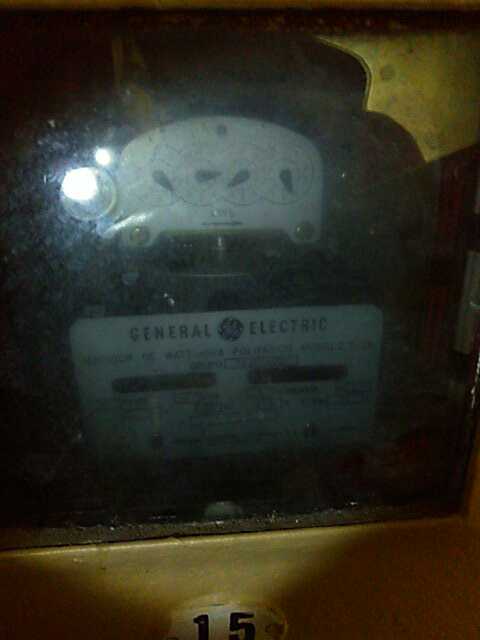}
    }
    \subfloat[distant capture]{
        \includegraphics[width=.18\linewidth]{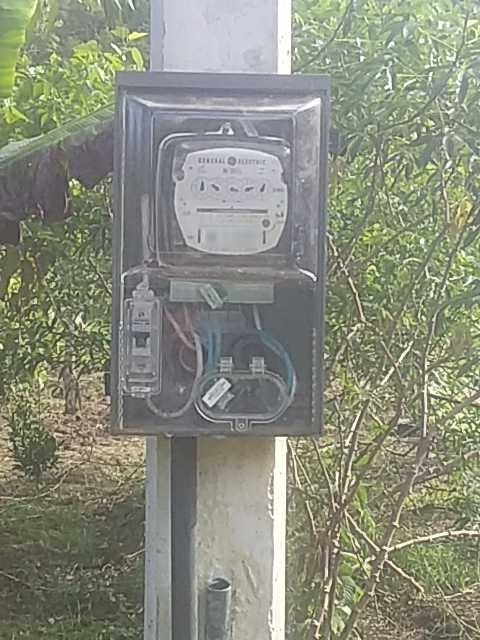}
    }
    \subfloat[reflections]{
        \includegraphics[width=.18\linewidth]{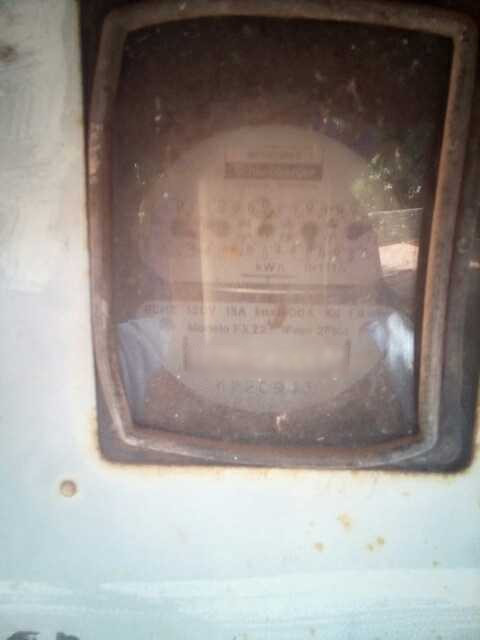}
    }
    
    \vspace{-1mm}

    \subfloat[dirt]{
        \includegraphics[width=.32\linewidth]{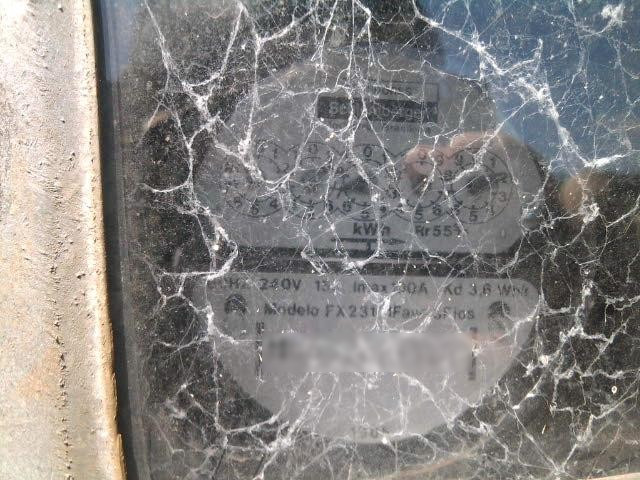}
    }  
    \subfloat[glare]{
        \includegraphics[width=.18\linewidth]{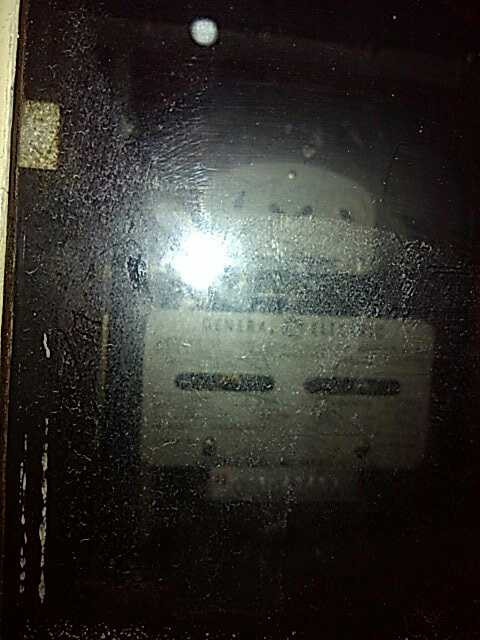}
    }      
    \subfloat[broken glass]{
        \includegraphics[width=.18\linewidth]{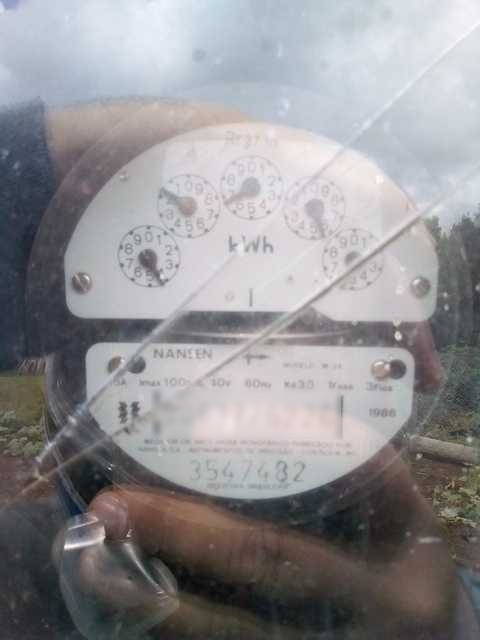}
    } 
    
    \caption{Samples of the challenging scenarios present on the provided images. We selected for the \dataset dataset $2{,}000$ images in which it was possible for a human to recognize the correct reading of the meter. We blurred the region containing the consumer unit number in each image due to privacy~constraints.
    }
    \label{fig:base-copel}
\end{figure*}

The main challenge of the proposed dataset is the quality of the images.
Low-end cameras, challenging environmental conditions and high compression are factors that have a high impact on the final image quality.
The challenging environmental conditions include: reflections, dirt, and broken glass, and low-quality acquisition may result in: noisy, blurred and low-contrast images. 
Fig.~\ref{fig:base-copel} illustrates the main image-quality issues described above.

In addition to the aforementioned quality issues, there are several types of meter templates and each manufacturer has its own dial model (with variations on the marks) and pointer design.
This variations combined with the image capture angle make it difficult to determine the exact pointed value.

Another challenge arises from the presence of clockwise and counter-clockwise dials -- for design purposes, each meter has alternating clock directions --, and the direction of the dials may differ depending on the meter model and~manufacturer.

\subsection{Evaluation Protocol and Metrics}

An evaluation protocol is necessary to enable fair comparison between different approaches. 
The dataset was randomly divided in three disjoints subsets: $1200$ images for training~($60$\%), $400$ images for validation~($20$\%) and the remaining $400$ images for testing~($20$\%).
Following recent works in which datasets were introduced~\cite{laroca2018robust,laroca2019convolutional,yang2019}, the subsets generated are explicitly available along with the \dataset~dataset.
 
To assess the recognition, three metrics are proposed: (i)~dial recognition rate, (ii)~meter recognition rate, and (iii)~mean absolute error.
As the main task is to correctly recognize the meter reading, which is a sequence of digits, the meter recognition rate consists of the comparison between the predicted sequence ($pred_m$) and the ground-truth sequence ($gt_m$), for each of the $N$ meters:
\begin{equation}
    MR_{rate} = \frac{1}{N}\sum_{m=1}^N match(pred_m,gt_m) 
\end{equation}

\[
    match(x,y)= 
\begin{cases}
    1,  & \text{if } x = y,\\
    0,  & \text{if } x \neq y.
\end{cases}
\]

For the dial recognition rate, we employed the \distance distance (also known as edit distance), a common measurement for computing distance between two sequences of characters. The \distance distance measures the minimum number of edits (addition, removal or replacement of characters) required to transform one sequence in the other.
\distance distance is suitable for our evaluation since it can handle small sequence errors in sequences that other metrics would treat as a big error.
For instance, if we have a ground-truth sequence $a=``1234"$ and a prediction sequence $b=``234"$, a per-character evaluation metric would consider the error equal to $4$, while \distance distance is equal to~$1$, as the difference between them is a single digit prediction.
The \distance distance between the sequences $a$ and $b$ can be determined using:
\begin{equation*}
lev_{a,b}(i,j)= 
        \begin{cases}
            max(i,j),  & if \ min(i,j)=0, \\
            lev'_{a,b}(i,j)     &\mbox{ otherwise,}
        \end{cases}
\end{equation*}
\noindent where:
\begin{equation*}
lev'_{a,b}(i,j) = min
\begin{cases}
lev_{a,b}(i-1,j)+1 \\
lev_{a,b}(i,j-1)+1 \\
lev_{a,b}(i-1,j-1) + 1_{(a_i \neq b_j).}
\end{cases}
\end{equation*}

The \distance distance between the prediction and the ground truth is computed and then divided by the longest sequence size between them.
This gives us the error.
Subtracting the error from $1$ gives us the dial recognition rate for each meter.
Finally, the mean of all recognition rates yields the total dial recognition error:
\begin{equation}
    DR_{rate} = \frac{1}{N} \sum_{m=1}^N \Big( 1 -   \frac{lev_{(pred_m,gt_m)}(|{}pred_m|{},|{}gt_m|{})}{max(|{}pred_m|{}, |{}gt_m|{})}  \Big)
\end{equation}

Considering that the sequence of digits that composes the meter reading is, in fact, a number (integer), correctly predicting the last digit in the sequence is not as important as correctly predicting the first one (i.e., the most significant digit).
In order to differentiate and penalize errors in the most significant digits, the mean absolute error is simple yet effective.
After converting the sequences to integers ($pred_m$ and $gt_m$ become the integers $p_m$ and $g_m$, respectively), the mean absolute error can be obtained~using:
\begin{equation}
    MA_{error} = \frac{1}{N} \sum_{m=1}^N |{}p_m - g_m|{}
\end{equation}

\section{Evaluated Approach}
\label{sec:approaches}

\begin{figure}[!htb]
    \centering
    \includegraphics[width=\linewidth]{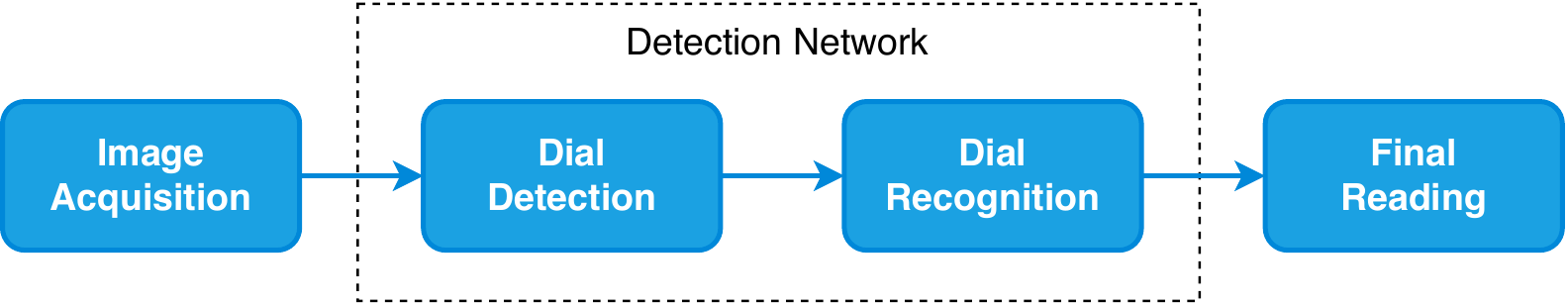}
    \caption{The main steps to perform dial meter reading.}
    \label{fig:pipeline}
\end{figure}

We chose two deep networks to evaluate: \faster and YOLO. 
The reason for treating dial meter reading as a detection problem arises from the previous successful approaches to~\gls{amr} using detection networks~\cite{laroca2019convolutional, fang2019}. 
\faster presented accurate results in several detection and recognition problems in the literature, while YOLO achieved reasonable results with a high rate of \gls*{fps}, improving the viability of mobile~applications. As illustrated in Fig.~\ref{fig:pipeline}, the proposed pipeline consists of (i)~image acquisition, (ii)~dial detection and recognition, and (iii)~final reading.

\subsection{Dial Detection}
We perform dial detection directly in the input images, that is, without first detecting the \gls*{roi}.
According to our experiments, presented in section~\ref{sec:results}, this approach achieves the highest F-score value. In other words, our recognition results are not significantly influenced by minor errors on the detection stage, making \gls*{roi} detection avoidable.

\subsection{Dial Recognition}
\faster is evaluated with the following residual networks as backbones replacing~VGG~\cite{simonyan2014very}: ResNet-50~\cite{he2016deep} (with 50 convolutional layers), ResNet-101~\cite{he2016deep} (with 101 convolutional layers) and ResNeXt-101~\cite{xie2017}.
According to~\cite{he2016deep}, ResNets outperform VGG and other several networks in classification tasks; therefore, they are used in our~experiments. 

For the YOLO-based models, we use the classifiers proposed along with the networks in~\cite{redmon2017yolo9000,redmon2018yolov3}.
YOLOv2 uses the Darknet-19 model as its backbone, which has $19$ convolutional layers (hence the name) and $5$ max-pooling layers.
YOLOv3, on the other hand, uses a network called Darknet-53 (with $53$ convolutional layers) for feature extraction; Darknet-53 can be seen as a hybrid approach between Darknet-19 and residual networks~\cite{redmon2018yolov3}.
We employed both YOLOv2 and YOLOv3 models in our~experiments in order to assess their speed/accuracy trade-off for this task.
 
\subsection{Final Reading}
The final reading is generated according to the position of the detected dial on the image (from leftmost to the rightmost dial).
Non-maximum suppression is performed using the \gls*{iou} metric ($IoU > 0.5$) and considering a maximum of $5$ dials per image, keeping only the dials predicted with higher confidence in order to avoid false~positives.
\section{Experimental Results}
\label{sec:results}

We evaluated the performance of the models based on YOLO and \faster to detect and recognize the dials simultaneously (note that we used pre-trained weights when fine-tuning both~networks). 
We performed our experiments on a CPU with a Quad-Core AMD Opteron 8387 2.8GHz processor, 64GB of RAM and an NVIDIA Titan Xp~GPU.
In order to stop the training process and select the best model for testing, we chose the \emph{\gls*{map}} evaluation metric, which has been commonly employed on object detection tasks~\cite{ren2015faster,redmon2016yolo,redmon2018yolov3}. 
The \gls*{map} can be calculated as follows:
\begin{equation}
    mAP = \frac{1}{c} \sum_{i=1}^c{AP_i} \, ,
\end{equation}
\noindent where $AP_i$ stands for the average precision value (for recall values from 0 to 1) of the $i$-th class. 

\subsection{Data Augmentation}
We generated new images by creating small variations to the training images to increase the generalization power of the networks.
Based on preliminary experiments carried out on the validation set, we generated seven times the number of training images (the combined number of original and augmented images was $9{,}600$).
The following transformations were randomly chosen for each image: random scaling [$-20$\%,~$20$\%], random translation [$-20$\%,~$20$\%], random rotation [$-15^{\circ}$,~$15^{\circ}$] and random shear [$-12$\%,~$12$\%].
The values, which are relative to the original size and position of the images, were chosen randomly within the defined~intervals.

\subsection{Evaluation}
First, we investigate the performance of the models in the dial detection task. 
The results are listed in Table~\ref{results-detection}.
For comparison, a common method proposed in the literature was evaluated: \gls*{hct}~\cite{zhang2016, zheng2017}. F-score was chosen as the evaluation metric, as it is often used to assess detection tasks.
As expected, deep learning-based methods (i.e., YOLO and \faster) outperformed \gls*{hct}, reaching very high F-score values.
\gls*{hct} did not cope well with the large variations on lighting, contrast and perspective found in our dataset~images.

\begin{table}[!ht]
 \centering
 
 \caption{Dial detection results achieved on the \dataset dataset.}
 
 \vspace{-1.5mm}
 \resizebox{.99\linewidth}{!}{
 \begin{tabular}{ccccc}
 \toprule
 \multirow{2}{*}{Detection Model} &
  \multirow{2}{*}{Backbone} & \multicolumn{3}{c}{(\%)}\\
  &  & Prec. & Recall & F-score\\ 
 \midrule
 \acrlong*{hct} & - & 53.27 & 55.28 & 54.25 \\
 Fast-YOLOv3 & Darknet  & 99.94 & 100.0 & 99.97\\
\textbf{YOLOv3} & \textbf{Darknet-53} & \textbf{100.0} & \textbf{100.0} & \textbf{100.0}\\
\faster & ResNet-50 & 100.0 & 99.94 & 99.97\\
\textbf{\faster} & \textbf{ResNet-101} & \textbf{100.0} & \textbf{100.0} & \textbf{100.0} \\
 \textbf{\faster} & \textbf{ResNeXt-101} & \textbf{100.0} & \textbf{100.0} & \textbf{100.0} \\ 
 \bottomrule
 \end{tabular}
 }
\label{results-detection}
\end{table}

We performed the recognition (reading) by combining the recognized digits (from the leftmost to the rightmost) and comparing them with the pointed values, using the metrics described in Section~\ref{sec:dataset}. 
The recognition results, as well as the \gls*{fps} rates obtained, are displayed in Table~\ref{results-recognition}. 

\begin{table}[!ht]
 \centering
 \caption{Recognition rate results obtained on the \dataset dataset.}
 
 \vspace{-1.5mm}
\resizebox{.99\linewidth}{!}{
 \begin{tabular}{cccccc}
 \toprule
\multirow{2}{*}{Method}  & \multirow{2}{*}{Input Size} & \multirow{2}{*}{\gls*{fps}} & \multicolumn{2}{c}{Recognition (\%) } & Mean Abs. \\
 &  &   & Dial & Meter & Error \\ 
\midrule
 Fast-YOLOv2 & $416\times416$ & $\textbf{244}$ & $79.61$ & $42.25$ & $5382.06$ \\ 
 Fast-YOLOv2 & $608\times608$ & $145$ & $85.24$ & $51.75$ & $3810.34$ \\ 
 \midrule
 Fast-YOLOv3 & $416\times416$ & $220$ & $83.27$ & $47.75$ & $6098.27$ \\ 
 Fast-YOLOv3  & $608\times608$ & $120$ & $86.60$ & $54.25$ &  $5183.82$ \\ 
 \midrule
 YOLOv2 & $416\times416$ & $67$ & $91.42$ & $68.00$ & $2615.23$ \\ 
 YOLOv2 & $608\times608$ & $40$ & $92.51$ & $71.25$ & $1924.98$\\ 
 \midrule
 YOLOv3 & $416\times416$ & $35$ & $93.00$ & $73.75$ & $1685.98$\\ 
 YOLOv3 & $608\times608$ & $20$ & $93.38$ & $74.75$ & $1591.16$\\ 
 \midrule
FR-CNN R-50 & $800\times800$ & $13$ & $92.56$ & $72.25$ & $1451.81$\\  
FR-CNN R-101 &  $800\times800$ & $11$ & $92.62$ & $71.75$ & $\textbf{1343.29}$\\ 
FR-CNN X-101 & $800\times800$ & $6$ & $\textbf{93.60}$ & $\textbf{75.25}$ & $1591.77$\\ 

\bottomrule
\end{tabular}
}
\label{results-recognition}
\end{table}

The best performing method was \faster (ResNext-101) followed by YOLOv3.
\faster obtained a $75.25$\% recognition rate per meter and $93.60$\% per dial, using $800\times800$-pixel images.
After YOLOv3, \faster with ResNet-101 performed better than ResNet-50 considering the recognition rate per dial.
Interestingly, ResNet-101 presented a lower hit rate considering the recognition at meter level. The lower hit rate is caused by the fact that ResNet-101 errors were better distributed across the images, while ResNet-50 concentrated the errors on fewer~images. 

The faster method was Fast-YOLOv2, using $416 \times 416$
images, achieving 244~\gls*{fps}.
Although YOLOv3 did not surpass \faster (ResNext-101) in recognition rates, the \gls*{fps} rates obtained were three times higher (20~\gls*{fps} and 6~\gls*{fps}, respectively).
Considering that the recognition rates achieved by YOLOv3 were not far behind, this model showed a promising trade-off between accuracy and~speed.

The best method regarding \emph{mean absolute error} was \faster (ResNet-101) with an error of $1343.29$. 
This means that the method's errors occurred less frequently (or were smaller) on the most significant digits.
Table~\ref{results-dist-errors} confirms this statement, as  \faster (ResNet-101) had fewer errors in the most significant dial (the leftmost).

\begin{table}[!ht]
 \centering
 \caption{Distribution of errors by dial position}
 
 \vspace{-1.5mm}

 \begin{tabular}{cccccc}
 \toprule
 \multirow{2}{*}{Dial Position} & \multicolumn{5}{c}{Frequency (\%)}   \\
 & 1 & 2 & 3 & 4 & 5 \\
\midrule
 YOLOv3 & $25.84$ & $15.44$ & $16.94$ & $25.99$ & $15.79$ \\
 FR-CNN (R-101) & $20.70$ & $18.66$ & $23.91$ & $19.83$ & $16.91$ \\
 FR-CNN (X-101) & $23.50$ & $18.55$ & $22.25$ & $18.04$ & $17.66$ \\

 \bottomrule

 \end{tabular}
\label{results-dist-errors}
\end{table}

\begin{figure}[!ht]
   \centering
   \includegraphics[width=\linewidth]{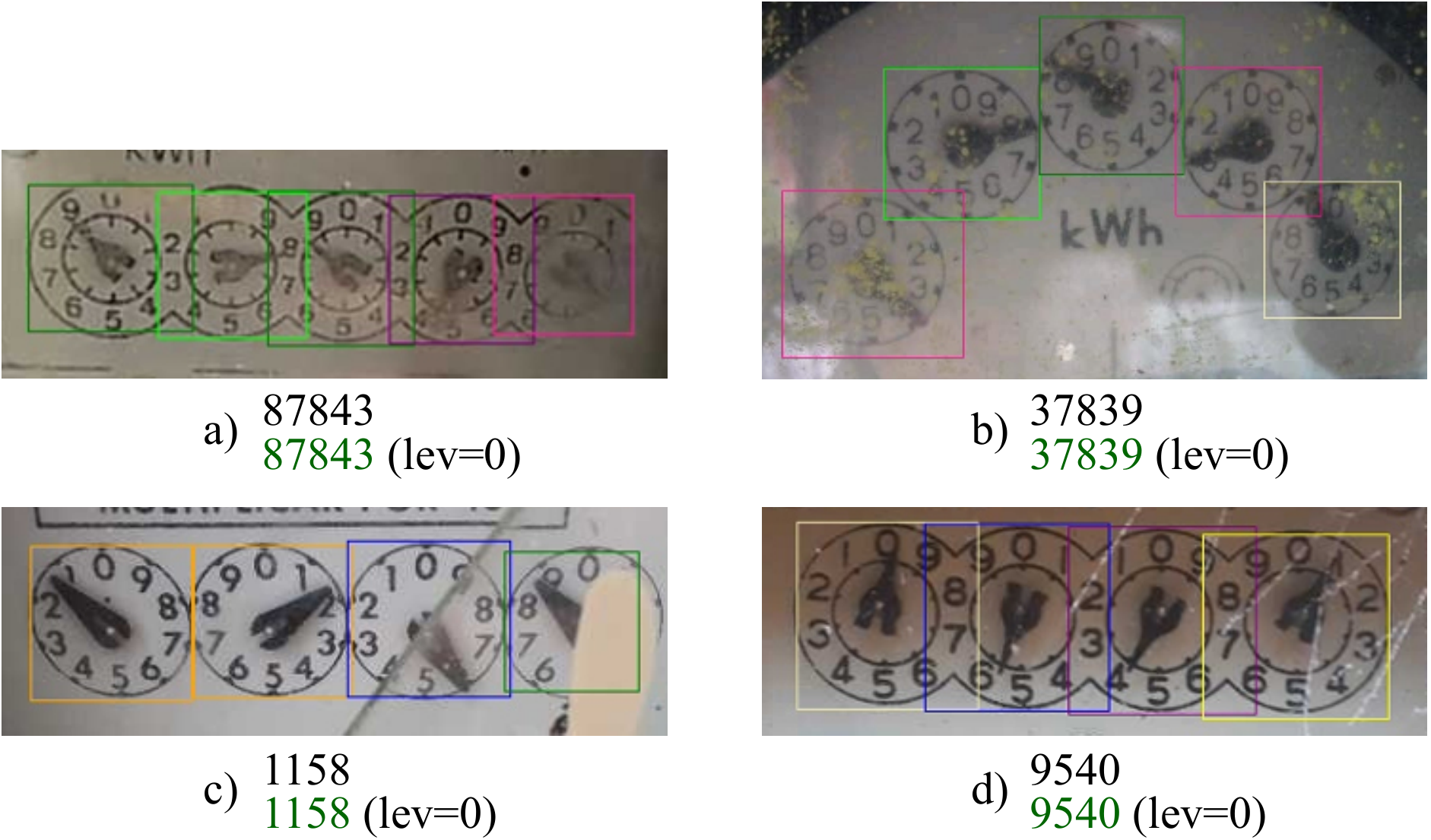}
   \vspace{-1.5mm}
    \caption{Ground-truth and prediction examples of correctly recognized meters, with their respective Levenshtein distance. 
    }
    \label{fig:only-correct}
\end{figure}

Fig.~\ref{fig:only-correct} presents some correct prediction results.
Note that the \distance distance between every correct prediction and its respective ground-truth annotation always equals~$0$.

\subsection{Error Analysis}
\begin{figure}[!ht]
   \centering
   \includegraphics[width=0.97\linewidth]{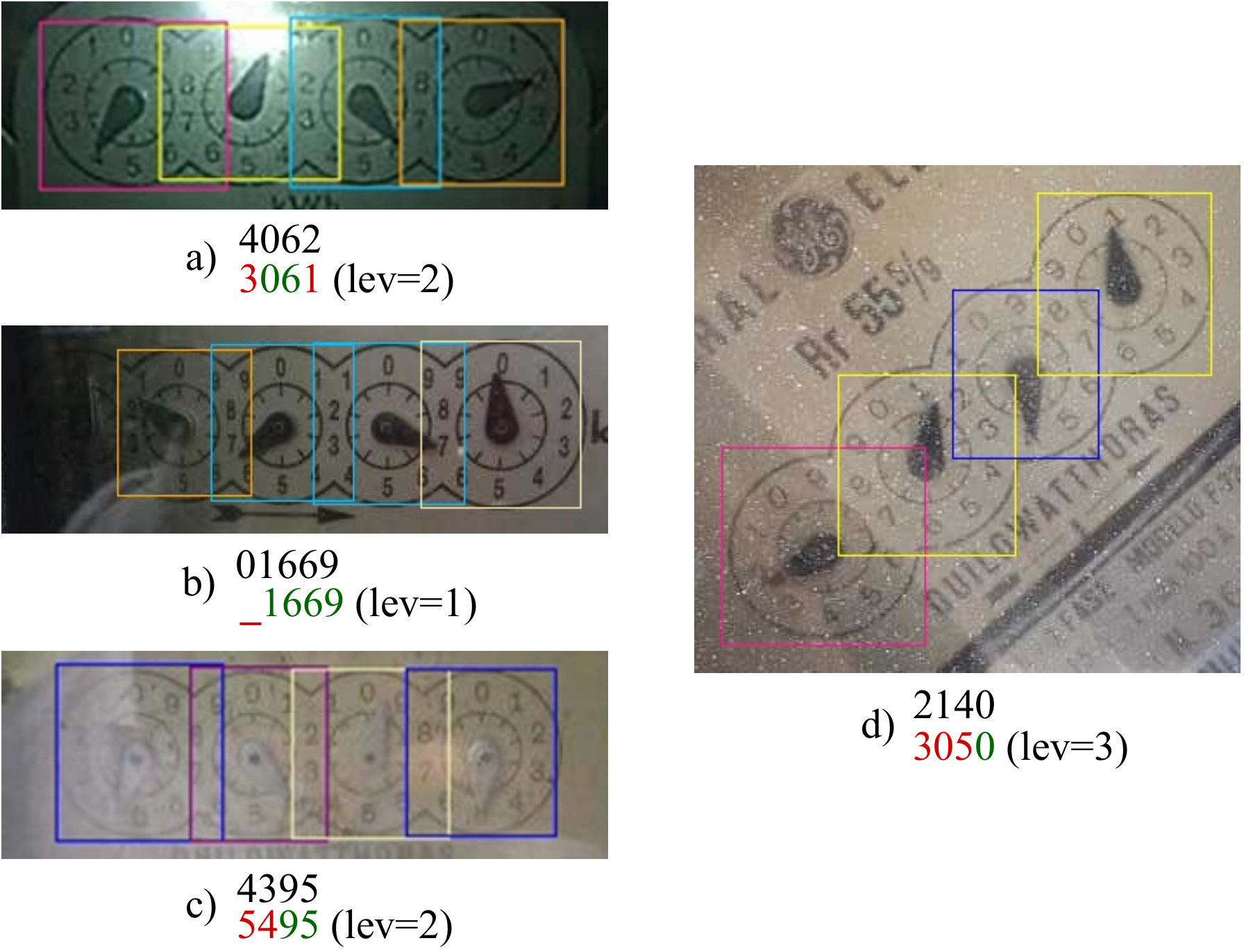}
   \vspace{-1.5mm}
    \caption{Ground-truth and prediction examples with their respective Levenshtein distance. The errors are marked in red and include: a) neighbor values, b) severe lighting conditions, c) neighbor value (second dial) and symmetry (first dial); and d) rotation.
    }
    \label{fig:only-errors}
\end{figure}

The most common errors in the presented approach are caused~by:
\begin{itemize}
    \item \textbf{Symmetry:} as there are clockwise and counterclockwise dials, when the digits are blurred, the method can not differentiate the direction and thus may output the mirrored value of the real prediction.
    \item \textbf{Neighbor value:} the most common error. Variables such as angle, lighting, shadows and occlusion (when the pointer is in front of the dial scale mark) can hinder the reading of a dial. Even between the authors, there were some disagreements regarding the correct pointed value in such~situations. 
    \item \textbf{Severe lighting conditions/Dirt:} shadows, glares, reflections and dirt may confuse the networks, especially in low-contrast images, where those artifacts may emerge more than the pointer, fooling the network to think that it is a pointer border, resulting in an incorrect prediction.
    \item \textbf{Rotation:} rotated images are harder to predict, as the pointed value is not in the usual position.
    The predictions may be assigned to neighbor digits that would be in the current angle of the pointer if the image was not rotated.

\end{itemize}
To illustrate all of the aforementioned causes of errors, some samples are presented in Fig.~\ref{fig:only-errors}.
Table~\ref{results-errors} summarizes the errors and their frequency on the best two methods: YOLOv3 and \faster (ResNeXt-101).
Note that most errors are caused by the neighbor values issue, when the pointer is in front of the mark, making it hard to determine if the pointed value is the one after or before the mark.

\begin{table}[!ht]
 \centering
 \caption{Type and frequency of errors obtained on evaluation.}
 
 \vspace{-1.5mm}

 \begin{tabular}{ccc}
 \toprule
 \multirow{2}{*}{Type of Error} & \multicolumn{2}{c}{Frequency}   \\
 & YOLOv3 & FR-CNN (X-101)  \\
\midrule
 Symmetry & $2$\% & $3$\% \\
 Neighbor value & $82$\% & $85$\% \\
 Lighting conditions / Dirt & $14$\% & $9$\% \\
 Rotation & $2$\% & $3$\% \\
 \bottomrule

 \end{tabular}
\label{results-errors}
\end{table}

\section{Conclusions}
\label{sec:conclusions}

Imaged-based \gls*{amr} is a faster and less laborious solution than manual on-site reading, and easier to deploy than the replacement of old~meters.
In this work, we presented the issues and challenges regarding the automatic reading of dial meters since there are many open challenges in this~context.

We introduced a public real-world dataset (shared upon request), called \dataset, for automatic dial meter reading, that includes $2{,}000$ fully annotated images acquired on site by employees of one of the largest companies of the Brazilian electricity sector~\cite{copel}.
As far as we know, this is the first public dataset containing images of dial meters.
The proposed dataset contains a well-defined evaluation protocol, which enables a fair comparison of different methods in future~works. 
Considering that the image scenario is challenging in most cases, the deep networks \faster and YOLO achieved promising results.
This straightforward approach, without \gls*{roi} detection or image preprocessing, simplified the traditional \gls*{amr} pipeline~\cite{vanetti2013gas,gallo2015robust,laroca2019convolutional}, reducing the number of steps required to obtain the dial meter~readings.

There is a lot of room for improvement, such as new methods to address the boundaries issues between the markers, which should solve most of the errors. 
In addition, a new loss function that penalizes errors on the leftmost dials should help to reduce the absolute error (minimizing the absolute error is of paramount importance to the service company).
\section*{Acknowledgments}
\label{sec:acknowledgments}
This work was supported by grants from the National Council for Scientific and Technological Development (CNPq), Grants \#313423/2017-2 and \#428333/2016-8) and the Coordination for the Improvement of Higher Education Personnel (CAPES) (Social Demand Program).
We gratefully acknowledge the support of NVIDIA Corporation with the donation of the Titan Xp GPU used for this research. We also thank the \acrfull*{copel} for providing the images for the \dataset~dataset.

\balance

\bibliographystyle{IEEEtran}
\bibliography{bib/bibtex}

\begin{thebibliography}{10}
\providecommand{\url}[1]{#1}
\csname url@samestyle\endcsname
\providecommand{\newblock}{\relax}
\providecommand{\bibinfo}[2]{#2}
\providecommand{\BIBentrySTDinterwordspacing}{\spaceskip=0pt\relax}
\providecommand{\BIBentryALTinterwordstretchfactor}{4}
\providecommand{\BIBentryALTinterwordspacing}{\spaceskip=\fontdimen2\font plus
\BIBentryALTinterwordstretchfactor\fontdimen3\font minus
  \fontdimen4\font\relax}
\providecommand{\BIBforeignlanguage}[2]{{%
\expandafter\ifx\csname l@#1\endcsname\relax
\typeout{** WARNING: IEEEtran.bst: No hyphenation pattern has been}%
\typeout{** loaded for the language `#1'. Using the pattern for}%
\typeout{** the default language instead.}%
\else
\language=\csname l@#1\endcsname
\fi
#2}}
\providecommand{\BIBdecl}{\relax}
\BIBdecl

\bibitem{vanetti2013gas}
M.~Vanetti, I.~Gallo, and A.~Nodari, ``Gas meter reading from real world images
  using a multi-net system,'' \emph{Pattern Recognition Letters}, vol.~34,
  no.~5, pp. 519--526, 2013.

\bibitem{gallo2015robust}
I.~Gallo, A.~Zamberletti, and L.~Noce, ``Robust angle invariant {GAS} meter
  reading,'' in \emph{International Conference on Digital Image Computing:
  Techniques and Applications}, Nov 2015, pp. 1--7.

\bibitem{li2019light}
C.~{Li}, Y.~{Su}, R.~{Yuan}, D.~{Chu}, and J.~{Zhu}, ``Light-weight spliced
  convolution network-based automatic water meter reading in smart city,''
  \emph{IEEE Access}, vol.~7, pp. 174\,359--174\,367, 2019.

\bibitem{laroca2019convolutional}
R.~{Laroca}, V.~{Barroso}, M.~A. {Diniz}, G.~R. {Gon{\c{c}}alves}, W.~R.
  {Schwartz}, and D.~{Menotti}, ``Convolutional neural networks for automatic
  meter reading,'' \emph{Journal of Electronic Imaging}, vol.~28, no.~1, p.
  013023, 2019.

\bibitem{b0}
\BIBentryALTinterwordspacing
U.~E.~I. Administration. (2019) Electric power annual 2018. [Online].
  Available: \url{https://www.eia.gov/electricity/annual/pdf/epa.pdf}
\BIBentrySTDinterwordspacing

\bibitem{kabalci2016survey}
Y.~Kabalci, ``A survey on smart metering and smart grid communication,''
  \emph{Renewable and Sustainable Energy Reviews}, vol.~57, pp. 302--318, 2016.

\bibitem{ausgrid}
\BIBentryALTinterwordspacing
Ausgrid. (2020) Types of meters. [Online]. Available:
  \url{https://www.ausgrid.com.au/Your-energy-use/Meters/Type-of-meters}
\BIBentrySTDinterwordspacing

\bibitem{callmepower}
\BIBentryALTinterwordspacing
Callmepower. (2020) Types of electricity meters. [Online]. Available:
  \url{https://callmepower.com/useful-information/electricity-meter-types}
\BIBentrySTDinterwordspacing

\bibitem{copel}
\BIBentryALTinterwordspacing
Copel. (2020) {Energy Company Of Paran\'a}. [Online]. Available:
  \url{http://www.copel.com/hpcopel/english/}
\BIBentrySTDinterwordspacing

\bibitem{liu2019}
Y.~Liu, J.~Liu, and Y.~Ke, ``A detection and recognition system of pointer
  meters in substations based on computer vision,'' \emph{Measurement}, vol.
  152, p. 107333, 2020.

\bibitem{zheng2017}
W.~{Zheng}, H.~{Yin}, A.~{Wang}, P.~{Fu}, and B.~{Liu}, ``Development of an
  automatic reading method and software for pointer instruments,'' in
  \emph{International Conference on Electronics Instrumentation Information
  Systems}, June 2017, pp. 1--6.

\bibitem{huang2019}
Y.~{Huang}, X.~{Dai}, and Q.~{Meng}, ``An automatic detection and recognition
  method for pointer-type meters in natural gas stations,'' in \emph{Chinese
  Control Conference}, July 2019, pp. 7866--7871.

\bibitem{jiale2011}
H.~Jiale, L.~En, T.~Bingjie, and L.~Ming, ``Reading recognition method of
  analog measuring instruments based on improved hough transform,'' in
  \emph{International Conference on Electronic Measurement Instruments},
  vol.~3, Aug 2011, pp. 337--340.

\bibitem{fang2019}
Y.~{Fang}, Y.~{Dai}, G.~{He}, and D.~{Qi}, ``A mask {RCNN} based automatic
  reading method for pointer meter,'' in \emph{Chinese Control Conference},
  July 2019, pp. 8466--8471.

\bibitem{tang2015}
Y.~{Tang}, C.~{Ten}, C.~{Wang}, and G.~{Parker}, ``Extraction of energy
  information from analog meters using image processing,'' \emph{IEEE
  Transactions on Smart Grid}, vol.~6, no.~4, pp. 2032--2040, July 2015.

\bibitem{vega2013}
R.~{Ocampo-Vega} \emph{et~al.}, ``Image processing for automatic reading of
  electro-mechanical utility meters,'' in \emph{Mexican International
  Conference on Artificial Intelligence}, Nov 2013, pp. 164--170.

\bibitem{gomez2018cutting}
L.~Gómez, M.~Rusiñol, and D.~Karatzas, ``Cutting sayre's knot: Reading scene
  text without segmentation. {A}pplication to utility meters,'' in \emph{IAPR
  Intern. Workshop on Document Analysis Systems}, 2018, pp. 97--102.

\bibitem{tsai2019}
C.~{Tsai}, T.~D. {Shou}, S.~{Chen}, and J.~{Hsieh}, ``Use {SSD} to detect the
  digital region in electricity meter,'' in \emph{International Conference on
  Machine Learning and Cybernetics~(ICMLC)}, July 2019, pp. 1--7.

\bibitem{yang2019}
F.~{Yang}, L.~{Jin}, S.~{Lai}, X.~{Gao}, and Z.~{Li}, ``Fully convolutional
  sequence recognition network for water meter number reading,'' \emph{IEEE
  Access}, vol.~7, pp. 11\,679--11\,687, 2019.

\bibitem{zhang2016}
L.~{Zhang} \emph{et~al.}, ``Pointer-type meter automatic reading from complex
  environment based on visual saliency,'' in \emph{International Conference on
  Wavelet Analysis and Pattern Recognition}, July 2016, pp. 264--269.

\bibitem{he2019}
P.~{He}, L.~{Zuo}, C.~{Zhang}, and Z.~{Zhang}, ``A value recognition algorithm
  for pointer meter based on improved {Mask-RCNN},'' in \emph{International
  Conference on Information Science and Technology}, 2019, pp. 108--113.

\bibitem{he2016deep}
K.~{He}, X.~{Zhang}, S.~{Ren}, and J.~{Sun}, ``Deep residual learning for image
  recognition,'' in \emph{IEEE Conference on Computer Vision and Pattern
  Recognition~(CVPR)}, June 2016, pp. 770--778.

\bibitem{simonyan2014very}
K.~Simonyan and A.~Zisserman, ``Very deep convolutional networks for
  large-scale image recognition,'' in \emph{International Conference on
  Learning Representations~(ICLR)}, 2015, pp. 1--12.

\bibitem{ren2015faster}
S.~{Ren} \emph{et~al.}, ``Faster {R-CNN}: Towards real-time object detection
  with region proposal networks,'' \emph{IEEE Transactions on Pattern Analysis
  and Machine Intelligence}, vol.~39, no.~6, pp. 1137--1149, 2017.

\bibitem{redmon2016yolo}
J.~{Redmon}, S.~{Divvala}, R.~{Girshick}, and A.~{Farhadi}, ``You only look
  once: Unified, real-time object detection,'' in \emph{IEEE Conference on
  Computer Vision and Pattern Recognition}, June 2016, pp. 779--788.

\bibitem{redmon2018yolov3}
J.~Redmon and A.~Farhadi, ``{YOLO}v3: An incremental improvement,'' \emph{arXiv
  preprint}, vol. arXiv:1804.02767, 2018.

\bibitem{nodari2011multi}
A.~Nodari and I.~Gallo, ``A multi-neural network approach to image detection
  and segmentation of gas meter counter.'' in \emph{IAPR Conference on Machine
  Vision Applications~(MVA)}, 2011, pp. 239--242.

\bibitem{liu2016ssd}
W.~Liu \emph{et~al.}, ``{SSD}: Single shot multibox detector,'' in
  \emph{European Conference on Computer Vision~(ECCV)}, 2016, pp. 21--37.

\bibitem{redmon2017yolo9000}
J.~Redmon and A.~Farhadi, ``{YOLO}9000: Better, faster, stronger,'' in
  \emph{IEEE Conference on Computer Vision and Pattern Recognition~(CVPR)},
  July 2017, pp. 6517--6525.

\bibitem{xie2017}
S.~Xie, R.~Girshick, P.~Dollar, Z.~Tu, and K.~He, ``Aggregated residual
  transformations for deep neural networks,'' in \emph{IEEE Conference on
  Computer Vision and Pattern Recognition~(CVPR)}, July 2017.

\bibitem{liu2019deep}
L.~Liu \emph{et~al.}, ``Deep learning for generic object detection: A survey,''
  \emph{International Journal of Computer Vision}, Oct 2019.

\bibitem{laroca2019efficient}
R.~{Laroca}, L.~A. {Zanlorensi}, G.~R. {Gon{\c{c}}alves}, E.~{Todt}, W.~R.
  {Schwartz}, and D.~{Menotti}, ``An efficient and layout-independent automatic
  license plate recognition system based on the {YOLO} detector,'' \emph{arXiv
  preprint}, vol. arXiv:1909.01754, pp. 1--14, 2019.

\bibitem{severo2018benchmark}
E.~{Severo}, R.~{Laroca}, C.~S. {Bezerra}, L.~A. {Zanlorensi},
  D.~{Weingaertner}, G.~{Moreira}, and D.~{Menotti}, ``A benchmark for iris
  location and a deep learning detector evaluation,'' in \emph{International
  Joint Conference on Neural Networks (IJCNN)}, July 2018, pp. 1--7.

\bibitem{laroca2018robust}
R.~{Laroca}, E.~{Severo}, L.~A. {Zanlorensi}, L.~S. {Oliveira}, G.~R.
  {Gon{\c{c}}alves}, W.~R. {Schwartz}, and D.~{Menotti}, ``A robust real-time
  automatic license plate recognition based on the {YOLO} detector,'' in
  \emph{International Joint Conference on Neural Networks (IJCNN)}, July 2018,
  pp. 1--10.

\end{thebibliography}

\end{document}